\documentclass[10pt,twocolumn,letterpaper]{article}

\usepackage{cvpr}              

\usepackage{multirow}
\usepackage{makecell} 
\usepackage{graphicx} %
\usepackage{subcaption}
\usepackage{threeparttable} 
\usepackage[accsupp]{axessibility}  

\definecolor{cvprblue}{rgb}{0.21,0.49,0.74}
\usepackage[pagebackref,breaklinks,colorlinks,allcolors=cvprblue]{hyperref}
\makeatletter
\def\thanks#1{\protected@xdef\@thanks{\@thanks
        \protect\footnotetext{#1}}}
\makeatother
\def\paperID{24640} 
\def\confName{CVPR}
\def\confYear{2026}

\title{Next-Scale Prediction: A Self-Supervised Approach \\ for Real-World Image Denoising}

\author{
Yiwen Shan$^1$ \and Haiyu Zhao$^1$ \and Peng Hu$^1$ \and Xi Peng$^{1,2}$ \and Yuanbiao Gou$^{1,\dag}$\thanks{$^{\dag}$ Corresponding author} 
\and $^1$College of Computer Science, Sichuan University, Chengdu, China\\
$^2$National Key Laboratory of Fundamental Algorithms and Models for Engineering \\ Numerical Simulation, Sichuan University, Chengdu, China\\
{\tt\small \{shan.yiwen.ml, haiyuzhao.gm, penghu.ml, pengx.gm, gouyuanbiao\}@gmail.com}
}

\begin{document}
\maketitle
\begin{abstract}
Self-supervised real-world image denoising remains a fundamental challenge, arising from the antagonistic trade-off between decorrelating spatially structured noise and preserving high-frequency details. Existing blind-spot network (BSN) methods rely on pixel-shuffle downsampling (PD) to decorrelate noise, but aggressive downsampling fragments fine structures, while milder downsampling fails to remove correlated noise. To address this, we introduce Next-Scale Prediction (NSP), a novel self-supervised paradigm that decouples noise decorrelation from detail preservation. NSP constructs cross-scale training pairs, where BSN takes low-resolution, fully decorrelated sub-images as input to predict high-resolution targets that retain fine details. As a by-product, NSP naturally supports super-resolution of noisy images without retraining or modification. Extensive experiments demonstrate that NSP achieves state-of-the-art self-supervised denoising performance on real-world benchmarks, significantly alleviating the long-standing conflict between noise decorrelation and detail preservation. The code is available at \url{https://github.com/XLearning-SCU/2026-CVPR-NSP}.
\end{abstract}

\section{Introduction}
\label{sec:intro}

Self-supervised image denoising~\cite{n2n,nbr2nbr,li2023spatially,s2s,blind2unblind,nr2n} aims to estimate the underlying noise-free images from their corrupted observations, without relying on any ground-truth supervision. The ill-posed nature and the absence of clean images make this task challenging and of broad practical applicability. 

A common paradigm for self-supervised image denoising is based on the blind-spot network (BSN)~\cite{n2v,n2s,laine,jang2023self,gao2023bs,chen2024exploring}, which predicts the clean value of a pixel from its surrounding pixels. This approach relies on the assumption that the noise is pixel-wise independent. If the noise exhibits spatial correlation, the network inevitably learns this correlation and predicts the noise in the target pixel from the nearby noisy pixels, which deviates from the denoising objective. Consequently, BSN-based methods are limited in handling real-world images, where noise introduced by Image Signal Processing~\cite{isp} pipeline often exhibits strong spatial correlation. As confirmed by prior studies~\cite{apbsn, sdap, tbsn, lgbpn}, BSN-based methods generally fail to achieve satisfactory results in real-world image denoising.

\begin{figure}[tb]
    \centering
    \includegraphics[width=\linewidth]{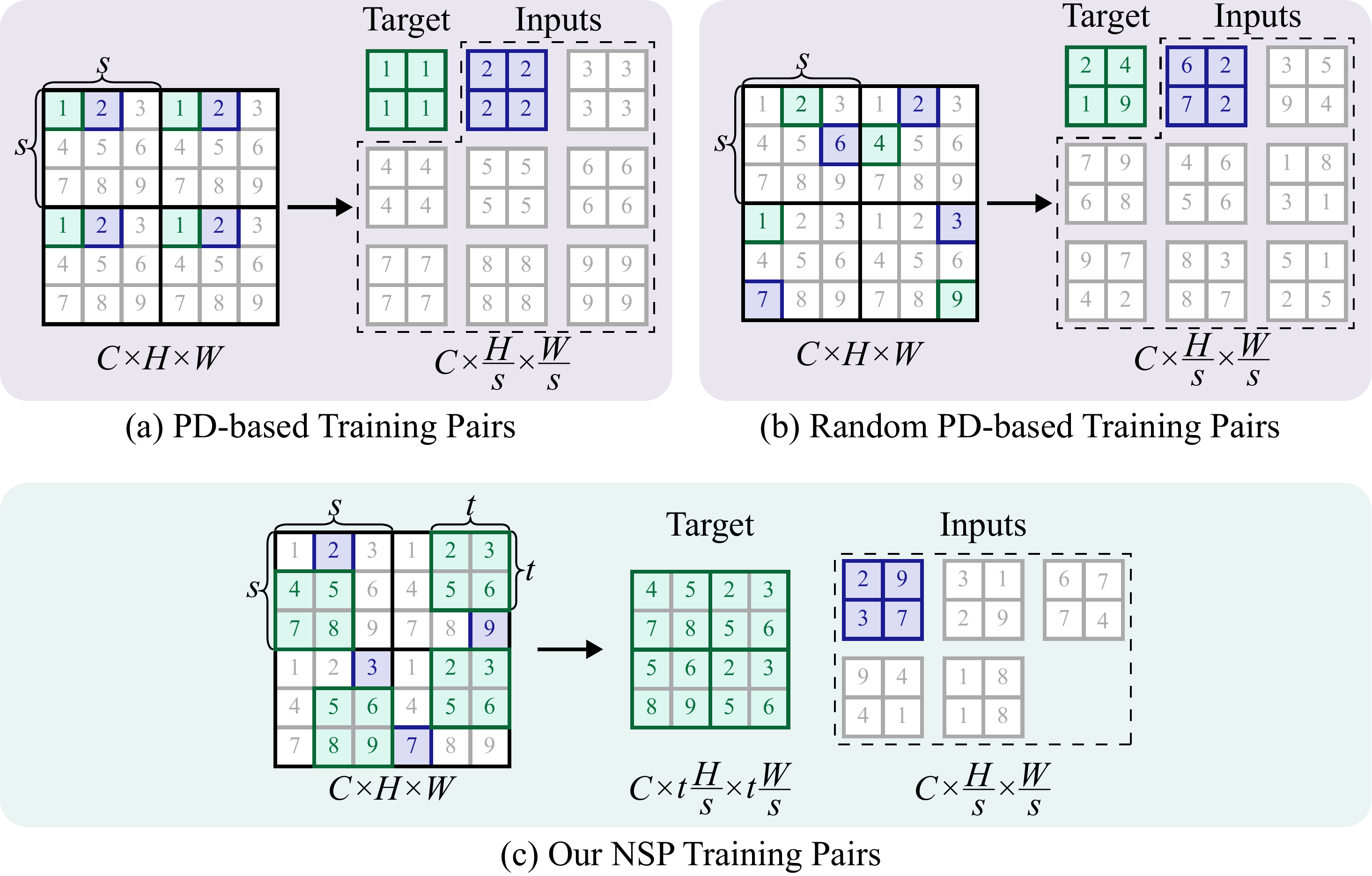}
    \caption{Illustration of different training-pair construction strategies for BSN. Inspired by visual autoregressive modeling, our NSP method constructs lower-scale inputs for noise decorrelation and high-scale targets for detail preservation.}
    \label{fig_PD_rPD}
\end{figure}

To enable BSN to handle real-world noise, researchers use pixel-shuffle downsampling (PD)~\cite{pdd}, which splits an image into many smaller sub-images. With a large downsampling factor, PD can effectively dismantle spatially correlated noise, turning it into nearly pixel-wise independent disturbances that BSN can cleanly remove. But that very aggressiveness comes at a steep price: it fragments the high-resolution structures, leaving the network to learn from tiny patches and crippling its ability to recover fine details. In some cases, BSN even misinterprets high-resolution structures as noise and removes them unintentionally~\cite{AliasingArtifact,apbsn}. Reduce the factor and detail returns, yet the noise regains spatial correlation that BSN cannot easily eliminate. In short, PD simultaneously undoes the correlation you must break and erases the resolution you must preserve. This antagonistic trade-off is subtle, fundamental, and central to making real-world denoising truly work.

This inherent contradiction motivates us to seek a more balanced strategy which decorrelates the noise more effectively while preserving the fine details. Ideally, such a strategy should retain the denoising advantages of PD while mitigating its destructive effect on high-resolution structures. Inspired by recent advances in Visual Autoregressive Modeling~\cite{var}, we propose Next-Scale Prediction (NSP) as a principled solution to this dilemma. Rather than relying on a fixed PD factor, NSP performs denoising through a coarse-to-fine, next-scale prediction process. Specifically, BSN takes sub-images produced with a large PD factor as input, and predicts their higher-resolution counterparts associated with a smaller PD factor. This design enables NSP to effectively decorrelate the spatially-correlated noise at lower scales while preserving the fine details at higher scales. By decoupling noise decorrelation from detail preservation, NSP alleviates the above intrinsic conflict present in PD-based BSN methods, providing a robust paradigm for real-world image denoising.

The contributions are summarized as follows:
\begin{itemize}
    \item \textbf{Next-Scale Prediction}: A new self-supervised paradigm for real-world image denoising that effectively addresses the conflict between noise decorrelation and detail preservation.
    \item \textbf{Data-Pair Construction}: A new strategy for constructing cross-scale/resolution training pairs that blocks access to noise-correlated pixels, coupled with a minimal BSN modification. 
    \item As a by-product of next-scale prediction, our method provides a feasible way to super-resolve real-world noisy images, without retraining or modifying the model.
\end{itemize}
\section{Related Works}
\label{sec:rel}
In this section, we review related work on self-supervised image denoising and visual autoregressive modeling.

\subsection{Self-Supervised Image Denoising}
Different from traditional image denoising~\cite{li2023spatial,he2020interactive,gou2020clearer,ffdnet,gou2022multi,tian2020deep}, which relies on noisy-clean pairs for training, self-supervised image denoising seeks to recover clean images directly from noisy observations. It leverages intrinsic image statistics rather than requiring paired supervision.

A foundational work of self-supervised image denoising is Noise2Noise~\cite{n2n}, which demonstrates that a denoiser can be trained using only corrupted image pairs. By replacing clean targets with another noisy observation of the same image, Noise2Noise achieves comparable performance to the models trained with clean targets. Furthermore, Noise2Void~\cite{n2v} and Noise2Self~\cite{n2s} proposed a blind-spot masking paradigm to generate pseudo noisy-noisy pairs from a single noise image. Specifically, these approaches mask a pixel and predict its clean value using surrounding noisy pixels. Since the masked pixels are excluded from the receptive field, they could be used as targets to supervise the training of the blind-spot network (BSN). However, this paradigm is inefficient since it trains the network using only masked pixels rather than the entire image at once. To address this, Laine et al.~\cite{laine} and DBSN~\cite{d-bsn} modified the network architecture by masking feature maps within the convolutions, enabling all pixels in the input image to serve as supervision signals for the BSN, thereby significantly enhancing training efficiency.

However, in real-world scenarios, the noise often exhibits strong spatial correlations, which pose challenges for these methods that rely on local correlations for restoration. To address this, several studies have explored strategies to break the spatial dependencies in the noise, enabling BSN to perform effective denoising in real-world scenarios. For example, AP-BSN~\cite{apbsn} introduces asymmetric pixel-shuffle downsampling for training and inference to break spatial noise correlation. SDAP~\cite{sdap} proposes random sub-sample generation strategy to disrupt noise dependencies during training. TBSN~\cite{tbsn} leverages transformer-based attention with masked spatial and grouped channel mechanisms to maintain the blind-spot constraint while capturing long-range dependencies.

Although these methods effectively reduce spatial noise correlations to enable real-world denoising, they also disrupt the natural correlations between image pixels, which could compromise image details. In contrast, we propose a novel paradigm for self-supervised image denoising that preserves fine-grained details. Specifically, it constructs training pairs using a downsampled low-resolution image to disrupt noise correlations and a high-resolution target to allow the network to learn fine-grained structures. With this design, our paradigm achieves effective denoising while minimally compromising image fidelity.

\subsection{Visual Autoregressive Modeling}

Visual Autoregressive Modeling (VAR)~\cite{var}, awarded the NeurIPS 2024 Best Paper, is a recently proposed and highly promising paradigm for image generation. Inspired by the human tendency to perceive and create images from coarse global structures to fine local details, VAR enables a transformer~\cite{transformer} to predict the next higher-scale token map conditioned on all token maps from previous scales. Compared to traditional next-token prediction, this next-scale prediction (NSP) is more natural for images, which expand in two-dimensional space rather than one-dimensional sequence. By avoiding the flattening of 2D into 1D, VAR preserves spatial structure and addresses the fundamental limitation of image autoregressive models. In practice, VAR significantly outperforms previous autoregressive baselines and surpasses strong diffusion models on multiple metrics.

Motivated by these observations, we propose NSP to disentangle noise decorrelation from detail preservation. The BSN first operates on a downsampled scale where noise is effectively decorrelated for removal, and subsequently predicts its high-resolution counterpart to restore fine details. This coarse-to-fine hierarchy ensures that denoising and structural reconstruction are addressed independently at their optimal scales.
\section{Proposed Method}

\begin{figure*}[tb]
    \centering
    \includegraphics[width=0.95\linewidth]{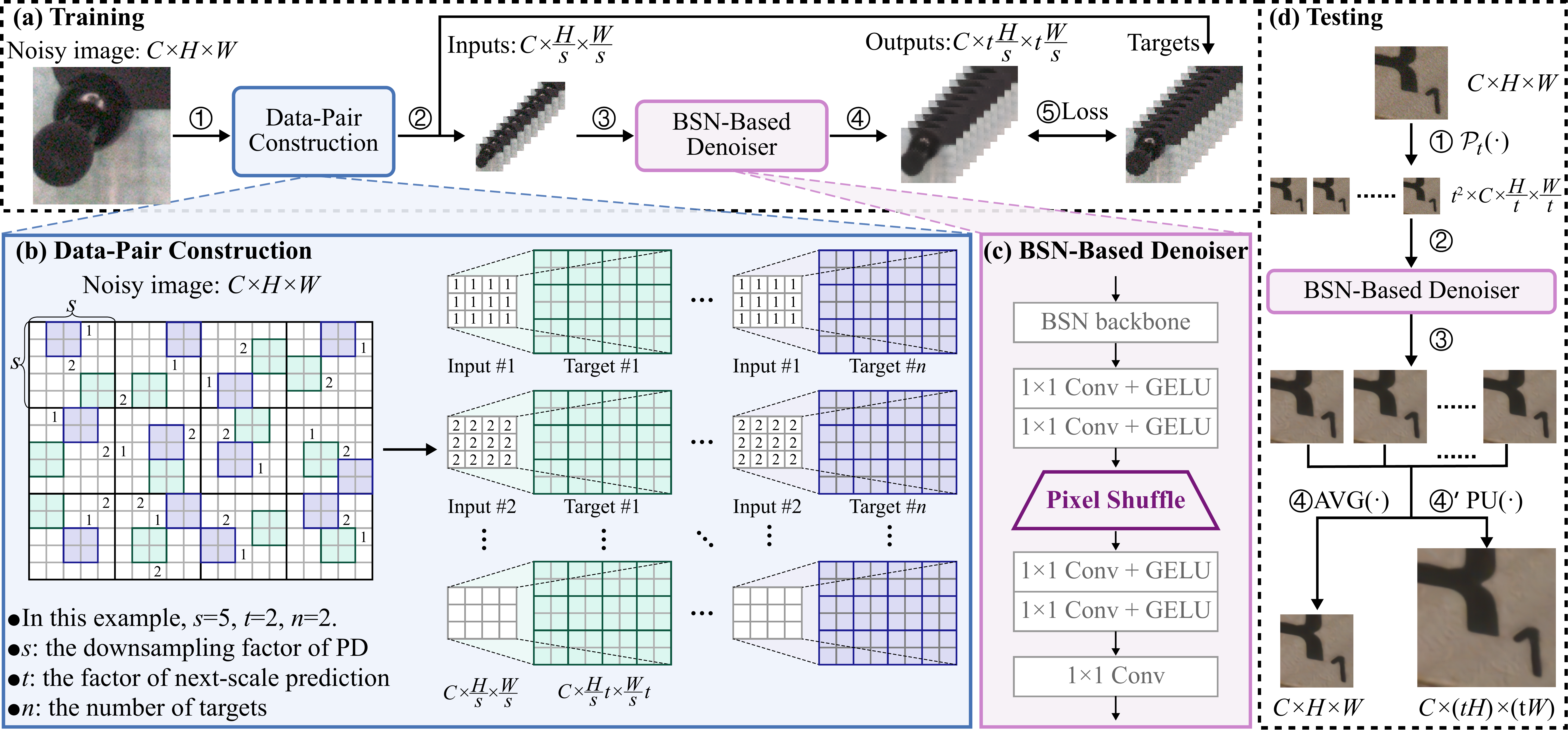}
    \caption{Framework of the proposed NSP. It is built on the PD-based BSN paradigm but reformulates denoising as a next-scale prediction task. The BSN first processes sub-images generated with a large PD factor and then learns to predict their higher-resolution counterparts corresponding to a smaller PD factor. In this way, NSP explicitly decouples the objectives of noise decorrelation and detail preservation.}
    \label{fig_framework}
\end{figure*}

In this section, we start with a brief review of the foundational denoising models, followed by a detailed explanation of our proposed method.

\subsection{PD-based BSN Methods}

BSN~\cite{n2v} is an effective architecture for self-supervised image denoising. By assuming zero-mean, pixel-wise independent noise, it predicts the clean value of each pixel from its surrounding noisy pixels. Formally, the training of BSN $f_{\theta}: \mathbb{R}^{C\times H\times W} \rightarrow \mathbb{R}^{C\times H\times W}$ can be formulated as
\begin{equation}
    \min_{\theta} \mathcal{L}\big( f_{\theta}(\mathbf{I}) - \mathbf{I} \big), 
    \label{eq_BSN}
\end{equation}
where $\theta$ is the parameter set, $\mathbf{I}$ is the noisy image, and $\mathcal{L}$ is certain loss function.

A fundamental limitation of BSN is that it can only remove pixel-wise independent noise, making it less effective for real-world images where noise often exhibits strong spatial correlation. To address this, PD~\cite{pdd, tbsn} decomposes a real-world noisy image into multiple sub-images (Fig.~\ref{fig_PD_rPD}(a)), effectively transforming spatially correlated noise into approximately pixel-wise independent noise. Building on this, later work~\cite{sdap} introduces randomness into the pixel-shuffle process. As shown in Fig.~\ref{fig_PD_rPD}(b), a single pixel is randomly selected from each $s \times s$ patch to form a sub-image. This substantially increases the number of possible sub-images, which can be calculated as $N = \sum_{i=0}^{s^2-1} (s^2 - i)^{\frac{HW}{s^2}}$, far exceeding $s^2$ sub-images generated by standard PD.

Despite notable advances, a fundamental bottleneck remains. To completely decorrelate spatially structured noise in real-world images, PD must use a large downsampling factor $s$, so that formerly neighboring, noise-correlated pixels are redistributed into different sub-images. That redistribution indeed weakens noise correlations, but it simultaneously destroys the high-frequency cues that encode fine structure. BSN trained on these sub-images lacks the statistical signal needed to recover subtle edges and textures. Reducing $s$ preserves those cues, yet it also preserves spatial noise correlation, leaving BSN with residual, signal-like noise that cannot distinguish from true detail.

In essence, PD-based BSN methods face an identifiability problem: the operations that decorrelate spatially structured noise inevitably remove the high-frequency cues that distinguish true image details from noise. This is not a mere tuning issue but an intrinsic trade-off: \textit{\textbf{noise decorrelation and detail preservation act on the same spatial dependencies in opposite directions, so any solution must explicitly decouple the two objectives rather than hope to meet them simultaneously}}.

\subsection{Framework Overview}

Motivated by above observation, we introduce Next-Scale Prediction (NSP), a self-supervised framework designed to explicitly decouple noise decorrelation and detail preservation. To be specific, NSP builds upon the PD-based BSN paradigm but reformulates denoising as a next-scale prediction problem. BSN first operates on sub-images produced with a large PD factor and then learns to predict their high-scale counterparts corresponding to a smaller PD factor. In this way, \textbf{\textit{noise removal is performed at the lower scale, where the noise has been largely decorrelated, while detail recovery occurs at the higher scale, where the detail has been largely preserved.}}

The overall framework of NSP is illustrated in Fig.~\ref{fig_framework}. Given a real-world noisy image $\mathbf{I} \in \mathbb{R}^{C\times H\times W}$, NSP first constructs cross-scale training pairs using our Data-Pair Construction strategy (Sec.~\ref{sec_pair_const}):
\begin{equation}
    (\mathbf{I}_s, \mathbf{I}_t) \;=\; \mathcal{G}(\mathbf{I}; s, t),
\end{equation}
where $\mathcal{G}$ denotes the proposed pair-construction operator. $s$ is the PD factor to generate the sub-image $\mathbf{I}_s \in \mathbb{R}^{C\times \frac{H}{s}\times \frac{H}{s}}$, and $t$ represents the relative scale of the target image $\mathbf{I}_t \in \mathbb{R}^{C\times \frac{H}{s}t\times \frac{H}{s}t}$ w.r.t. $\mathbf{I}_s$. In this formulation, $\mathbf{I}_s$ serves as the coarse, noise-decorrelated input, whereas $\mathbf{I}_t$ provides the high-scale reference that retains more spatial structure and detail.

With the cross-scale training pairs $(\mathbf{I}_s, \mathbf{I}_t)$, BSN $\mathcal{F}_\theta$ is trained to predict the high-scale target $\mathbf{I}_t$ from the coarse-scale input $\mathbf{I}_s$:
\begin{equation}
    \hat{\mathbf{I}}_t = \mathcal{F}_\theta(\mathbf{I}_s),
\end{equation}
where $\theta$ denotes the parameter set, which is optimized by minimizing the reconstruction loss between the prediction and target:
\begin{equation}
    \mathcal{L}(\theta) = \big\| \hat{\mathbf{I}}_t - \mathbf{I}_t \big\|_1.
\end{equation}
Since both $\mathbf{I}_s$ and $\mathbf{I}_t$ are derived from the same noisy image $\mathbf{I}$, this training process forms a self-supervised loop:
\begin{equation}
    \mathbf{I} \xrightarrow{\mathcal{G}(\cdot; s, t)} (\mathbf{I}_s, \mathbf{I}_t) \xrightarrow{\mathcal{F}_\theta} (\hat{\mathbf{I}}_t, \mathbf{I}_t) \xrightarrow{\mathcal{L}(\theta)} \text{update } \theta.
\end{equation}
Through this next-scale prediction formulation, BSN learns to perform noise removal at the coarse scale, where noise is largely decorrelated, while progressively refining spatial details guided by the high-scale targets.

Given a test noisy image $\mathbf{I'}\in\mathbb{R}^{C\times H\times W}$, let $\mathcal{P}_t$ denote PD operator that first decomposes $\mathbf{I'}$ into $t^2$ sub-images:
\begin{equation}
    \{\mathbf{I'}^{(i)}\}_{i=1}^{t^2} \;=\; \mathcal{P}_t(\mathbf{I'}), 
    \quad \mathbf{I'}^{(i)}\in\mathbb{R}^{C \times \frac{H}{t} \times \frac{W}{t}}.
\end{equation}
Then, each sub-image is independently processed by the trained BSN $\mathcal{F}_\theta$:
\begin{equation}
    \hat{\mathbf{I'}}^{(i)} \;=\; \mathcal{F}_\theta\big(\mathbf{I'}^{(i)}\big), \quad i=1,\dots,t^2,
\end{equation}
where $\hat{\mathbf{I'}}^{(i)}\in\mathbb{R}^{C\times H\times W}$ denotes the denoised output corresponding to the $i$-th sub-image at the high scale. Finally, the denoised image is obtained by averaging these $t^2$ reconstructed results:
\begin{equation}
    \hat{\mathbf{I'}} \;=\; \mathrm{AVG}(\{\hat{\mathbf{I'}}^{(i)}\}_{i=1}^{t^2}) \in \mathbb{R}^{C\times H\times W}.
\end{equation}

As a by-product, our pipeline also yields a feasible way to produce a $t\times$ super-resolved image of the noisy input without retraining. Let PU denote the pixel-shuffle upsampling that reassembles $t^2$ denoised sub-images into a single image of size $tH \times tW$. The resulting super-resolved output is then given by
\begin{equation}
    \hat{\mathbf{I}}^{\uparrow t} \;=\; \text{PU} \big(\{\hat{\mathbf{I'}}^{(i)}\}_{i=1}^{t^2}\big) \in \mathbb{R}^{C\times (tH)\times (tW)}.
\end{equation}

\subsection{Data-Pair Construction}
\label{sec_pair_const}

To construct effective cross-scale training pairs, we adhere to three key principles:
\begin{itemize}
    \item \textbf{Blocking noise-correlated pixels across scales} to guarantee BSN learns true noise removal rather than exploiting residual noise correlations. 
    \item \textbf{Maintaining structural consistency} to preserve the pixels spatial arrangement in the high-scale targets, ensuring better reconstruction of details.
    \item \textbf{Leveraging random sampling} to generate a wide variety of training pairs, covering diverse noise realizations and structural patterns.
\end{itemize}
Following these principles, our cross-scale pair construction process is illustrated in Fig.~\ref{fig_framework}(b). 

Given a real-world noisy image $\mathbf{I} \in \mathbb{R}^{C \times H \times W}$, we first divide it into $s \times s$ non-overlapping patches:
\begin{equation}
\{\mathbf{P}_{i,j}\} = \text{Patch}(\mathbf{I}; s), \quad i / j=1,\dots,\frac{H}{s}/\frac{W}{s},
\end{equation}
where $s$ denotes both the patch size and the PD factor from the original image to the sub-images.

Next, for each patch $\mathbf{P}_{i,j} \in \mathbb{R}^{C \times s \times s}$, we perform random sampling to select $t \times t$ pixels that construct the high-scale target:
\begin{equation}
\mathbf{L}_{i,j} = \text{Sample}(\mathbf{P}_{i,j}; t) \in \mathbb{R}^{C \times t \times t},
\label{eq_high_sample}
\end{equation}
where $t \in (1, s)$ is the scaling factor of the cross-scale pair. 

The remaining $(s^2 - t^2)$ pixels in $\mathbf{P}_{i,j}$ are then distributed across $(s^2 - t^2)$ sub-images:
\begin{equation}
\{\mathbf{I}_{i,j}^{(k)}\}_{k=1}^{s^2-t^2} = \text{Distribute}(\mathbf{P}_{i,j} \setminus \mathbf{L}_{i,j}),
\label{eq_low_sample}
\end{equation}
where each $\mathbf{I}_{i,j}^{(k)} \in \mathbb{R}^{C \times 1 \times 1}$ corresponds to a pixel position.

Finally, each sub-image is paired with the same high-scale target, producing the final cross-scale training set:
\begin{equation}
\{(\mathbf{I}_{i,j}^{(k)}, \mathbf{L}_{i,j})\}_{k=1}^{s^2-t^2}.
\label{eq_data_pairs}
\end{equation}

\begin{figure}[tb]
    \centering
    \includegraphics[width=\linewidth]{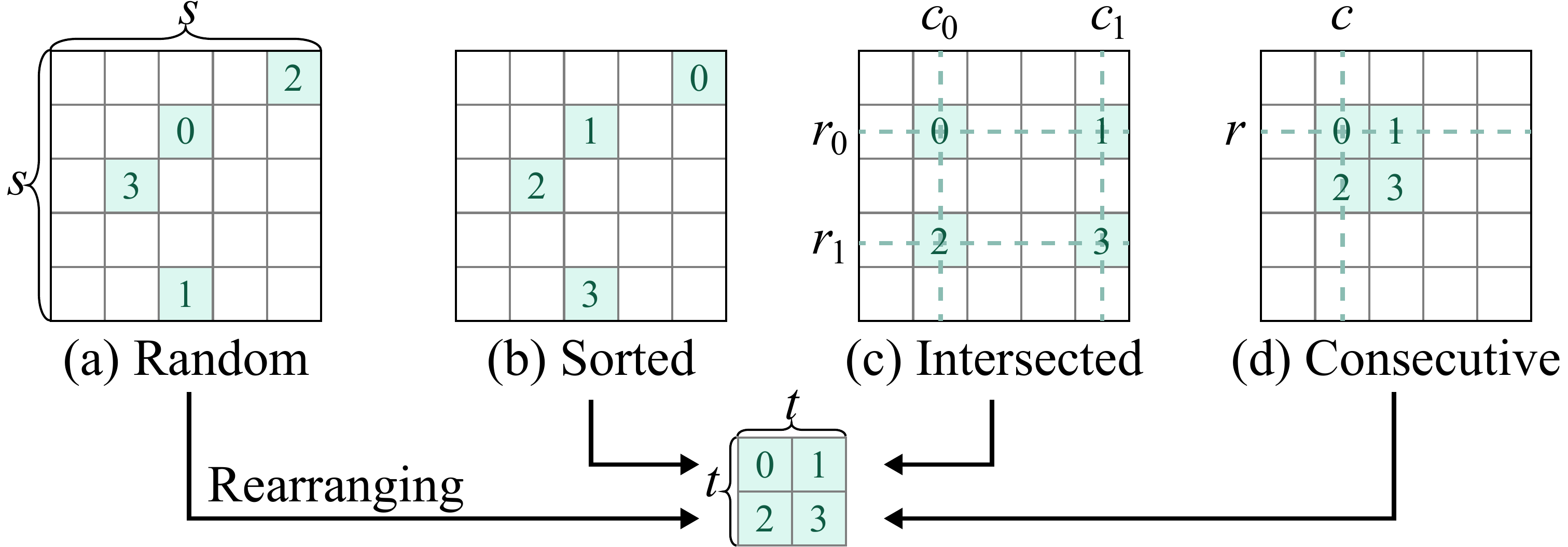}
    \caption{Alternative strategies for high-scale target construction. (a) Randomly select $t^2$ pixels. (b) Randomly select $t^2$ pixels and sort in a row-major order. (c) Select the pixels at the intersections of $t$ random rows and columns. (d) Select a consecutive $t\times t$ patch. The comparison between those strategies can be found in Tab.~\ref{tab_strtegy_n}.}
    \label{fig_target_select}
\end{figure}

For the Eq.~(\ref{eq_high_sample}), we design several alternative sampling strategies, as illustrated in Fig.~\ref{fig_target_select}, and empirically find that those preserving the relative spatial arrangement of pixels within each patch perform the best. Consequently, we employ the consecutive patch strategy as our default choice, which can produce $(s-t+1)^2$ possible targets.

For the Eq.~(\ref{eq_low_sample}), this one-to-one random distribution assigns noise-correlated pixels to different sub-images, which can generate $\sum_{i=0}^{s^2-t^2-1} (s^2-t^2-i)^\frac{HW}{s^2}$ sub-images. Consequently, BSN takes these decorrelated sub-images as inputs to predict high-scale targets composed of distinct pixels, thereby effectively blocking noise correlations across scales. 

For the Eq.~(\ref{eq_data_pairs}), while the formulation constructs a single high-scale target, the strategy naturally generalizes to the multi-target case. Multiple targets are generated by repeatedly sampling $t \times t$ pixels not occupied by previous targets, while the remaining pixels are assigned to sub-images. Let the number of targets be $n \in [1, \lfloor s^2 / t^2 \rfloor]$, so that the number of inputs is $(s^2 - n \cdot t^2)$. As a result, the training pairs are formed as the Cartesian product of the inputs and targets, yielding a total of $n \cdot (s^2 - n \cdot t^2)$ pairs per image.

\subsection{BSN-Based Denoiser}
\label{sec_denoiser}

Within the NSP framework, a BSN is required to predict the next higher-scale sub-images rather than operating within the same scale as in its original design. To achieve this, a pixel shuffle layer is integrated into BSN to facilitate scale transformation. To keep the blind-spot property, the layer is inserted between the $1 \times 1$ convolutional blocks near the network tail, as shown in Fig.~\ref{fig_framework}(c).

\section{Experiments}

\begin{table*}[t]
\caption{Quantitative comparisons on real-world noisy image datasets. \dag denotes results from the previous SIDD evaluation server, which is no longer accessible.}
\label{tab_de}
\centering
\resizebox{0.95\linewidth}{!}{\begin{tabular}{lllcccc}
\toprule
\multirow{2}{*}{\textbf{Type}} & \multirow{2}{*}{\textbf{Train Data}} & \multirow{2}{*}{\textbf{Method}} & \multirow{2}{*}{\textbf{\#Param}} & \textbf{SIDD Validation} & \textbf{SIDD Benchmark} & \textbf{DND} \\
& & & & \textbf{PSNR/SSIM} & \textbf{PSNR/SSIM} & \textbf{PSNR/SSIM}\\
\midrule
\multirow{2}{*}{Non-Learning} & \multirow{2}{*}{None} 
& BM3D~\cite{bm3d} & - & 31.75/0.7061 & 34.26/0.6950\dag & 34.51/0.8507 \\&
& WNNM~\cite{wnnm} & - & 26.31/0.5240 & 30.52/0.4498\dag & 34.67/0.8646 \\ 
\midrule
\multirow{6}{*}{Supervised} 
&  & 
DnCNN~\cite{dncnn} & 0.66M & 26.20/0.4414 & 30.31/0.4371\dag & 32.43/0.7900 \\
&  & 
TNRD~\cite{tnrd} & - & 26.99/0.7440 & - & 33.65/0.8306 \\
& Paired & 
CBDNet~\cite{cbdnet} & 6.79M & 33.07/0.8655 & 34.51/0.8402 & 38.00/0.9400 \\ 
& Noisy-Clean & 
PDD~\cite{pdd} & 0.71M & 33.96/0.8195 & 35.22/0.8221 & 38.40/0.9434 \\ 
&  & 
RIDNet~\cite{ridnet} & 1.50M & 38.71/0.9511 & - & 39.25/0.9528 \\
&  & 
VDNet~\cite{vdnet} & 7.82M & 39.29/0.9109 & 39.49/0.9117\dag & 39.38/0.9518 \\
&  & 
DeamNet~\cite{deamnet} & 2.23M & 39.40/0.9169 & 39.58/0.9118\dag & 39.63/0.9531 \\
\midrule 
& Unpaired & GCBD~\cite{gcbd} & - & - & - & 35.58/0.9217 \\ 
& Noisy-Clean & D-BSN~\cite{d-bsn} & 6.62M & - & - & 37.93/0.9373 \\ 
Pseudo- & &
C2N~\cite{c2n} & 217.26M & 35.36/0.8901 & 36.06/0.8825\dag & 37.28/0.9237 \\ 
\cmidrule{2-7}
Supervised & Paired &
\multirow{2}{*}{R2R~\cite{r2r}} & \multirow{2}{*}{0.56M} & \multirow{2}{*}{35.04/0.8440} & 
\multirow{2}{*}{35.50/0.8550\dag} & \multirow{2}{*}{37.61/0.9368} \\ & Noisy-Noisy & & & & \\
\midrule
& & N2V~\cite{n2v} & 2.58M & 29.07/0.5915 & 31.77/0.5979 & 33.37/0.8412 \\ 
& & N2S~\cite{n2s} & 2.58M & 30.72/0.7870 & 32.57/0.6768 & 33.63/0.8564 \\ 
& & NEI2NEI~\cite{nbr2nbr} & 1.26M & 28.00/0.5890 & - & 31.40/0.7880 \\ 
 &  & 
CVF-SID~\cite{cvfsid} & 1.19M & 34.14/0.8550 & 35.03/0.8561 & 36.50/0.9233 \\ 
Self & Only Noisy & 
AP-BSN~\cite{apbsn} & 3.66M & 34.46/0.8296 & 36.09/0.8310 & 37.46/0.9244 \\
-Supervised&  & SDAP~\cite{sdap} & 3.66M & 36.58/0.8630 & 36.80/0.8529 & 37.71/0.9278 \\
&  & TBSN~\cite{tbsn} & 12.74M & 36.59/0.8574 & 37.08/0.8519 & \textbf{37.90}/0.9288 \\
&  & \textbf{NSP(DBSN)} & 3.75M & \textbf{37.02/0.8865} & \textbf{37.42/0.8748} & 37.80/\textbf{0.9319} \\
&  & \textbf{NSP(TBSN)} & 12.77M & \textbf{37.12/0.8853} & \textbf{37.46/0.8744} & 37.87/\textbf{0.9342} \\
\bottomrule
\end{tabular}}
\end{table*}

In this section, we conduct experiments to evaluate the proposed method. In the following, we first introduce the experimental settings, followed by quantitative and qualitative results on image denoising and super-resolution tasks. Finally, we perform analysis experiments to validate the effectiveness of our key designs.

\subsection{Experimental Settings}

To comprehensively evaluate the proposed NSP framework, we adopt two representative BSNs as backbones, namely DBSN and TBSN, which are CNN-based and Transformer-based architectures, respectively. In experiments, we denote these two variants as NSP(DBSN) and NSP(TBSN). The PD factor $s$, which determines the downsampling from the original image to the sub-images, is empirically set to 5, while the scaling factor $t$ from the sub-images to the high-scale targets is set to 2. Following prior works~\cite{apbsn,sdap,tbsn}, we set the patch size to $160 \times 160$ and the batch size to 16. Training is conducted for 750 epochs, with each epoch consisting of 400 iterations. The learning rate is fixed at 1e-4, and the optimizer is Adam with default parameters. All experiments are implemented in PyTorch on NVIDIA GeForce RTX 3090 and A800 GPUs.

\subsection{Experimental Results}

The denoising methods based on the proposed paradigm are trained on the noisy subset of SIDD Medium, which consists of 320 high-resolution real-world noisy images. For fairness, all comparison methods are evaluated on SIDD Validation, SIDD Benchmark and DND dataset. Specifically, both SIDD Validation and SIDD Benchmark have 1280 noisy images of size 256$\times$256, while DND contains 1000 noisy images of size 512$\times$512. 

\begin{figure*}[tb]
    \centering
    \subfloat[Noisy]{
        \includegraphics[width=0.14\textwidth]{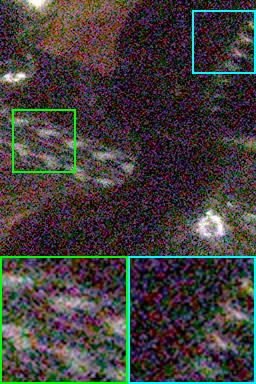}
    }
    \subfloat[CBDNet]{
        \includegraphics[width=0.14\textwidth]{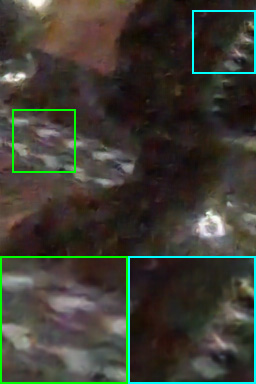}
    }
    \subfloat[PDD]{
        \includegraphics[width=0.14\textwidth]{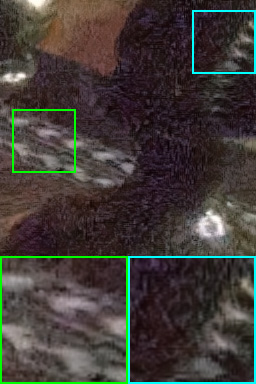}
    }
    \subfloat[RIDNet]{
        \includegraphics[width=0.14\textwidth]{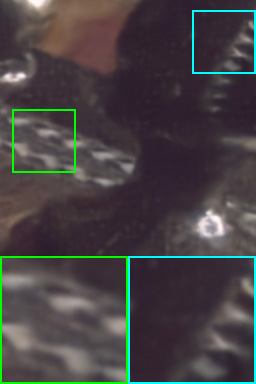}
    }
    \subfloat[N2V]{
        \includegraphics[width=0.14\textwidth]{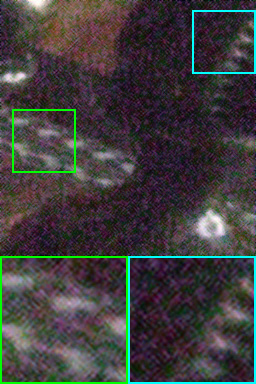}
    }
    \subfloat[CVF-SID]{
        \includegraphics[width=0.14\textwidth]{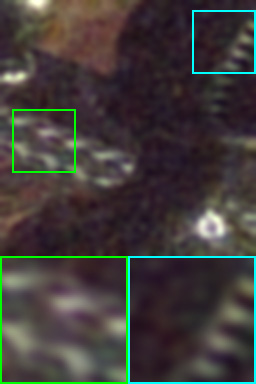}
    }
    \\
    \subfloat[AP-BSN]{
        \includegraphics[width=0.14\textwidth]{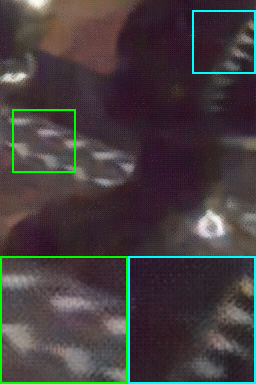}
    }
    \subfloat[SDAP]{
        \includegraphics[width=0.14\textwidth]{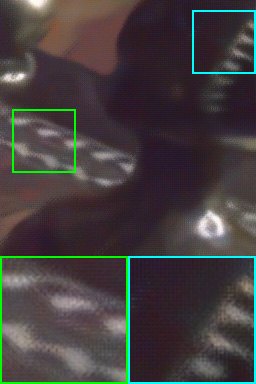}
    }
    \subfloat[TBSN]{
        \includegraphics[width=0.14\textwidth]{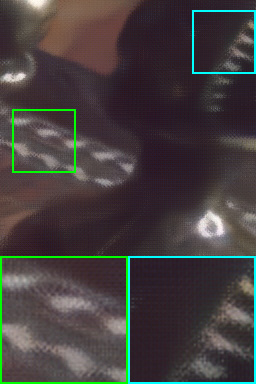}
    }
    \subfloat[NSP(DBSN)]{
        \includegraphics[width=0.14\textwidth]{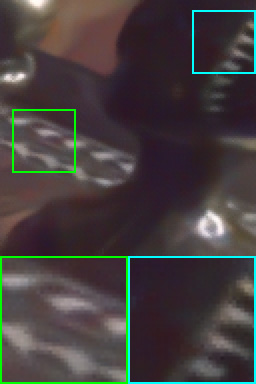}
    }
    \subfloat[NSP(TBSN)]{
        \includegraphics[width=0.14\textwidth]{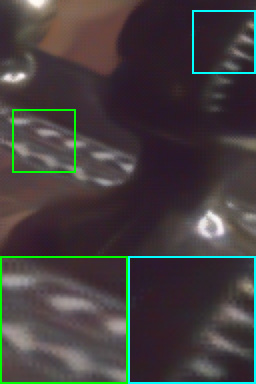}
    }
    \subfloat[GT]{
        \includegraphics[width=0.14\textwidth]{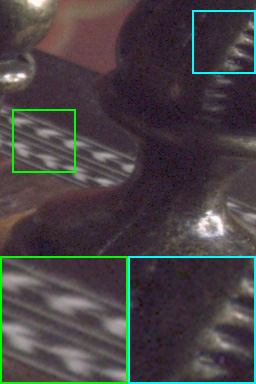}
    }
    \caption{Qualitative comparisons of image denoising on SIDD Validation. }
    \label{fig_DE_19_29}
\end{figure*}

The quantitative results of all comparison methods are presented in Table~\ref{tab_de}. As can be seen, NSP(DBSN) and NSP(TBSN) almost achieve the best performance among all self-supervised methods, except for the PSNR on DND dataset being slightly lower than TBSN by 0.03dB. Specifically, both NSP(DBSN) and SDAP use DBSN as the backbone with similar parameter amounts (3.75M and 3.66M). However, NSP(DBSN) outperforms SDAP by 0.44dB/0.0235 on SIDD Validation, according to the PSNR/SSIM metric. A similar result can be found between NSP(TBSN) and TBSN, both of which use the TBSN as the backbone. Such improvement achieved by the NSP paradigm is attributed to the better prediction of the high-scale details, as shown in the highlighted windows of Fig.~\ref{fig_DE_19_29}(h)$\sim$(j). Moreover, the performance of NSP(DBSN) and NSP(TBSN) are competitive with that of multiple supervised and pseudo-supervised methods. 
Among them, NSP(DBSN) outperforms DnCNN, TNRD, CBNDet, PDD, GCBD and C2N with at least 1.66dB margin in PSNR.

The qualitative results are shown in Fig.~\ref{fig_DE_19_29}. In Fig.~\ref{fig_DE_19_29}(e), Noise2Void (N2V) leaves too much noise in the denoised result. This verifies the BSN cannot remove the spatially-correlated noise effectively, since it would learn the spatial correlation inadvertently and predict the noise in each pixel. In Fig.~\ref{fig_DE_19_29}(g)$\sim$(i), the ``PD+BSN''-based counterparts, \ie, AP-BSN, SDAP, and TBSN, generate chessboard artifacts, which is due to no module can help the BSN learn to predict the higher-scale details from the lower-scale sub-images output by PD. By contrast, the proposed NSP paradigm compensates this by supervising the BSN to predict the higher-scale counterparts where more fine details are preserved. Hence, as shown in Fig.~\ref{fig_DE_19_29}(j)$\sim$(k), more details are recovered by NSP(DBSN) and NSP(TBSN). 

An interesting by-product of the proposed paradigm is noisy image super-resolution (SR), since the BSN predicts a higher-resolution version of the input. Different from most existing SR methods which focus solely on SR, the proposed paradigm performs the image denoising and SR simultaneously. To conduct the noisy image SR experiment, we first downsample the noisy subset of the SIDD Medium and SIDD Validation datasets using a factor of 2 and bicubic interpolation. Then, the proposed NSP(DBSN) is trained and tested on the corresponding downsampled datasets. All the settings are kept the same as those in the denoising experiment, except that the patch size is changed to 180$\times$180. The high-resolution results are obtained by the procedure along the right branch in Fig.~\ref{fig_framework}(d). 

For a fair comparison, all competing methods are trained on the downsampled noisy subset of SIDD Medium. To provide denoising capability for the SR methods, we first train the standard SDAP and then use its denoised outputs as the training inputs for the SR methods. Four zero-shot SR methods~\cite{zssr, rzsr, mzsr, dualsr} are selected and re-trained on the dataset to create comparable dataset-trained versions for evaluation.

\begin{figure}[tb]
    \centering
    \!\!\subfloat[Noisy LR input]{
        \includegraphics[width=0.24\linewidth]{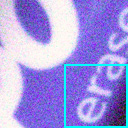}
    }
    \!\!\subfloat[ZSSR]{
        \includegraphics[width=0.24\linewidth]{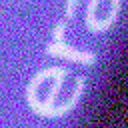}
    }
    \!\!\subfloat[SDAP+ZSSR]{
        \includegraphics[width=0.24\linewidth]{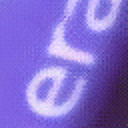}
    }
    \!\!\subfloat[SDAP+MZSR]{
        \includegraphics[width=0.24\linewidth]{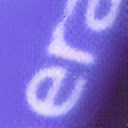}
    }

    \!\!\subfloat[SDAP+dualSR]{
        \includegraphics[width=0.24\linewidth]{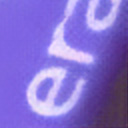}
    }
    \!\!\subfloat[SDAP+U-ZSSR]{
        \includegraphics[width=0.24\linewidth]{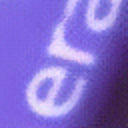}
    }
    \!\!\subfloat[NSP(DBSN)]{
        \includegraphics[width=0.24\linewidth]{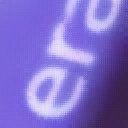}
    }
    \!\!\subfloat[GT]{
        \includegraphics[width=0.24\linewidth]{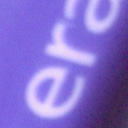}
    }
    \caption{Qualitative comparison of image denoising and super-resolution on SIDD Validation.}
    \label{fig_SR_10_29}
\end{figure}

The quantitative results are listed in Table~\ref{tab_sr}. In the first group, the zero-shot SR methods do not obtain satisfactory results, since they have no capacity of denoising and can only output the noisy high-resolution images, as shown in Fig. \ref{fig_SR_10_29}(b). In the following two groups, the two-stage ``SDAP+SR'' methods are able to output the denoised high-resolution results. Hence their PSNR and SSIM are significantly higher than those in the first group. In spite of this, the proposed NSP(DBSN) still achieves the highest results with a competitive amount of parameters. 
The qualitative results are shown in Fig. \ref{fig_SR_10_29}. Due to the difficulty of noisy image SR problem, the comparison methods generate the artifacts to some extent. In spite of this, the proposed NSP(DBSN) still recovers the edges more clearly and generates fewer artifacts than the two-stage methods.

\begin{table}[t]
\caption{Quantitative results on noisy image super-resolution over the SIDD Validation dataset. “SDAP+” denotes a pipeline that first applies SDAP for denoising and then performs super-resolution. “U-” denotes the variant trained on the dataset, as opposed to zero-shot training.}
\label{tab_sr}
\centering
\resizebox{0.8\linewidth}{!}{
\begin{tabular}{lccc}
    \toprule
    \textbf{Method} & \textbf{PSNR/SSIM} & \textbf{\#Param} \\
    \midrule
    ZSSR~\cite{zssr} & 25.38/0.4183 & 0.22M\\
    RZSR~\cite{rzsr} & 25.00/0.3982 & 3.04M\\
    MZSR~\cite{mzsr} & 25.93/0.4426 & 0.46M\\
    dualSR~\cite{dualsr} & 24.98/0.4473 & 0.41M\\
    \midrule
    SDAP+ZSSR & 34.39/0.8165 & 3.88M\\
    SDAP+RZSR & 34.43/0.8189 & 6.70M\\
    SDAP+MZSR & 34.17/0.8284 & 4.12M\\
    SDAP+dualSR& 34.67/0.8380 & 4.06M\\
    \midrule 
    SDAP+U-ZSSR & 35.10/0.8373 & 3.88M\\
    SDAP+U-RZSR$^{1}$ & - & 6.70M\\
    SDAP+U-MZSR$^{1}$ & - & 4.12M\\
    SDAP+U-dualSR$^{2}$ & 23.05/0.6123 & 4.06M\\
    \midrule 
    NSP(DBSN) & 35.19/0.8490  & \multirow{2}{*}{3.75M}\\
    NSP(DBSN)$^{3}$& \textbf{35.53/0.8771} & \\
    \bottomrule
\end{tabular}
}
\begin{tablenotes}
\footnotesize
\item $^{1}$ The design of the method precludes it from being adapted from a zero-shot setting to a standard training regime.
\item $^{2}$ Despite extensive attempts, training on a dataset consistently fails.
\item $^{3}$ At test time, directly input noisy LR image and output SR version without PD/PU.
\end{tablenotes}
\end{table}

\subsection{Analysis Experiments}
\label{sec_exp_analysis}

In this section, we test the influence of three important settings, \ie, the strategy of constructing the targets, the number of targets constructed from single image, and the upsampling factor from the sub-images to the targets. 

\textbf{Target Construction Strategies.}
To construct a high-scale target, $t^2$ pixels are first selected from each $s \times s$ patch and then rearranged into a $t \times t$ patch, as illustrated in Fig.~\ref{fig_framework}(b). During this rearrangement, it is important to preserve the relative positions of the selected pixels, as these positions encode structural information of the image. The degree of structural preservation depends on the selection strategy: purely random selection, such as in Fig.~\ref{fig_target_select}(a), often fails to maintain the original relative positions. Therefore, the choice of pixel selection strategy is critical. In this work, we design four alternative strategies shown in Fig.~\ref{fig_target_select}, where the purely random strategy serves as a baseline to evaluate the effectiveness of the other structured approaches.

Table~\ref{tab_strtegy_n} compares the performance of different target construction strategies. As shown in the first column, the purely random selection performs the worst, since the relative positions of the selected pixels are mostly lost. The second strategy, which sorts the randomly selected pixels in row-major order (Fig.~\ref{fig_target_select}(b)), improves performance by partially recovering the relative positions between some pixels. For instance, the relative order of pixels 1 and 2 is corrected compared to Fig.~\ref{fig_target_select}(a). However, this strategy still fails to preserve all relative positions; for example, the horizontal order between pixel 0 and 3 remains reversed.

\begin{table}[tb]
    \centering
    \caption{The comparison of different strategies of target construction and different number of targets $n$. The four strategies in Fig.~\ref{fig_target_select} are denoted as ``Random'', ``Sorted'', ``Intersected'' and ``Consecutive'', respectively.}
    \resizebox{\linewidth}{!}{
    \begin{tabular}{p{0.1cm}<{\centering}p{0.1cm}<{\centering}|p{1.55cm}<{\centering}p{1.55cm}<{\centering}p{1.55cm}<{\centering}p{1.5cm}<{\centering}}
        \toprule
        & & \!\!\!Random & \!\!\!Sorted & \!\!\!Intersected & \!\!\!Consecutive\\
        \!\!\!\!$n$ & \!\!\!\!\!\!\!Pairs & \!\!\!PSNR/SSIM & \!\!\!PSNR/SSIM & \!\!\!PSNR/SSIM & \!\!\!PSNR/SSIM\\
        \hline
        \!\!\!\!1 & \!\!\!21 & \!\!\!36.77/0.8835 & \!\!\!36.91/0.8835 & \!\!\!37.01/0.8854& \!\!\!\textbf{37.02/0.8865}\\
        \!\!\!\!2 & \!\!\!34 & \!\!\!36.87/0.8849 & \!\!\!36.97/0.8842 & \!\!\!37.02/0.8864& \!\!\!\textbf{37.11/0.8884}\\
        \!\!\!\!3 & \!\!\!39 & \!\!\!36.83/0.8837 & \!\!\!37.05/0.8848 & \!\!\!37.07/0.8855& \!\!\!\textbf{37.08/0.8861}\\
        \!\!\!\!4 & \!\!\!36 & \!\!\!36.89/0.8841 & \!\!\!37.01/0.8848 & \!\!\!\textbf{37.25/0.8865}& \!\!\!37.04/0.8855\\
        \!\!\!\!5 & \!\!\!25 & \!\!\!36.88/0.8841 & \!\!\!36.88/0.8827 & \!\!\!- & \!\!\!- \\
        \bottomrule
    \end{tabular}}
    \label{tab_strtegy_n}
\end{table}

To address this limitation, two additional strategies are proposed, as illustrated in Fig.~\ref{fig_target_select}(c) and (d). The third strategy randomly selects $t$ rows and $t$ columns and chooses the pixels at their intersections, while the fourth strategy selects a consecutive $t \times t$ patch. Both strategies perfectly preserve the relative positions of the selected pixels, leading to superior performance compared with the sorting-based strategy, as shown in the last two columns of Table~\ref{tab_strtegy_n}. Overall, these results highlight the critical importance of maintaining relative pixel positions in constructing high-scale targets.

\textbf{Number of Targets.}
The number of targets $n$ serves as a hyper-parameter that controls how many training pairs can be constructed from a single noisy image. As discussed in Section~\ref{sec_pair_const}, the total number of pairs is given by a quadratic function of $n$: $-t^2\!\cdot\!n^2 + s^2\!\cdot\!n$. While increasing $n$ can provide more training pairs and potentially improve the performance, it also incurs higher GPU memory consumption.

Given $s=5$ and $t=2$, the impact of the number of targets $n$ on denoising performance is summarized in the first few rows of Table~\ref{tab_strtegy_n}. When $n$ increases from 1 to 2, the total number of training pairs grows from 21 to 34, leading to consistent performance improvements across all target construction strategies. Moreover, the highest results for each strategy are obtained when the number of pairs exceeds 30, confirming that a larger number of pairs benefits the training of the proposed paradigm. Although the best performance is achieved with $n>1$, we report the results for $n=1$ in Table~\ref{tab_de} to ensure compatibility with widely available GPUs having 24GB of memory.

\textbf{Influence of Scaling Factor.}
The factor $t$ is a hyper-parameter controlling the prediction scale. When $t$ is too large, the mapping from low-resolution inputs to high-resolution targets becomes overly complex, as the targets are $t$ times larger than the inputs, making it harder for the denoiser to learn and leading to degraded performance. Conversely, if $t$ is too small, e.g., $t=1$, the paradigm degenerates to the conventional PD-based BSN method, which may fail to accurately recover fine details. Furthermore, at test time, $t$ determines the input size of the denoiser (Fig.~\ref{fig_framework}(d)). For $t=1$, the input is not downsampled, so the spatial correlation of noise remains, preventing the BSN from effectively removing it.

Those observations are confirmed by the results in Table~\ref{tab_ids}. The NSP paradigm achieves the best performance when $t=2$, while the performance degrades noticeably for both smaller and larger values of $t$. And, this trend holds consistently across different target construction strategies.

\begin{table}[tb]
    \centering
    \caption{The comparison of different upsampling factor $t$. The four strategies in Fig.~\ref{fig_target_select} are denoted as ``Random'', ``Sorted'', ``Intersected'' and ``Consecutive'', respectively.}
    \resizebox{0.9\linewidth}{!}{
    \begin{tabular}{p{0.01cm}<{\centering}|p{1.6cm}<{\centering}p{1.6cm}<{\centering}p{1.6cm}<{\centering}p{1.6cm}<{\centering}}
        \toprule
            & \!\!Random & \!\!Sorted & \!\!Intersected & \!\!Consecutive\\
        \!\!$t$ & \!\!PSNR/SSIM & \!\!PSNR/SSIM & \!\!PSNR/SSIM & \!\!PSNR/SSIM\\
        \hline
        \!\!1 & \!\!34.48/0.8415 & \!\!3415/0.8440 & \!\!33.97/0.8353 & \!\!34.12/0.8346 \\
        \!\!2 & \!\!\textbf{36.77/0.8835} & \!\!\textbf{36.91/0.8835} & \!\!\textbf{37.01/0.8854} & \!\!\textbf{37.02/0.8865}\\
        \!\!4 & \!\!35.87/0.8615 & \!\!36.29/0.8682 & \!\!36.56/0.8716 & \!\!36.46/0.8694\\
        \bottomrule
    \end{tabular}}
    \label{tab_ids}
\end{table}

\section{Conclusion}

In this paper, we propose Next-Scale Prediction (NSP), a self-supervised paradigm for real-world image denoising. NSP enables a BSN to denoise low-resolution sub-images, where noise correlations are largely broken, while simultaneously predicting their high-resolution counterparts to preserve the fine details. This coarse-to-fine approach explicitly decouples the objectives of noise decorrelation and detail preservation, resolving the intrinsic conflict inherent in conventional PD-based BSN methods. To facilitate training, we introduce a data-pair construction strategy that generates a diverse set of cross-scale pairs from a single noisy image. Extensive experiments on real-world benchmarks validate the effectiveness of NSP, and comprehensive analyses investigate the influence of key hyper-parameters and design choices.

\section*{Acknowledgments}

This work was supported in part by the Postdoctoral Fellowship Program ( Grade C ) of China Postdoctoral Science Foundation under Grant GZC20251052; in part by the Fundamental Research Funds for the Central Universities under Grant CJ202303; in part by Sichuan Science and Technology Planning Project under Grant 2024NSFTD0038; in part by Fundamental and Interdisciplinary Disciplines Breakthrough Plan of the Ministry of Education of China under Grant JYB2025XDXM610.
{
    \small
    \bibliographystyle{ieeenat_fullname}
    \bibliography{main}
}

\end{document}